\documentclass[letterpaper,10pt,conference]{ieeeconf}    
\IEEEoverridecommandlockouts                             
\usepackage{graphicx}
\usepackage{verbatim}                                                    
\usepackage[hyphens]{url}

\title{\LARGE \bf
CIDI-Lung-Seg: A Single-Click Annotation Tool for Automatic Delineation of Lungs from CT Scans
}

\author{Awais Mansoor$^{1}$, Ulas Bagci$^{*,1}$, Brent Foster$^{1}$, Ziyue Xu$^{1}$, Deborah Douglas$^{2}$,\\ Jeffrey M. Solomon$^{2}$, Jayaram K. Udupa$^{3}$, and Daniel J. Mollura$^{1}$
\thanks{*This work is supported by Center for Infectious Disease Imaging (CIDI), National Institutes of Health (NIH), Bethesda MD 20892.}
\thanks{$^{1}$A. Mansoor, U. Bagci, B. Foster, Z. Xu, and D. Mollura are with the Department of Radiology and Imaging Sciences, National Institutes of Health (NIH), Bethesda MD 20892.}
\thanks{$^{2}$D. Douglas and J. Solomon are with the Department of Radiology and Imaging Sciences, The NIAID Integrated Research Facility, National Institutes of Health (NIH), Fredrick MD 21702.}
\thanks{$^{3}$ J. Udupa is with the Department of Radiology, University of Pennsylvania, Philadelphia PA 19104.}%
}

\begin{document}

\sloppy

\maketitle
\thispagestyle{empty}
\pagestyle{empty}

\begin{abstract}

Accurate and fast extraction of lung volumes from computed tomography (CT) scans remains in a great demand in the clinical environment because the available methods fail to provide a generic solution due to wide anatomical variations of lungs and existence of pathologies. Manual annotation, current gold standard, is time consuming and often subject to human bias. On the other hand, current state-of-the-art fully automated lung segmentation methods fail to make their way into the clinical practice due to their inability to efficiently incorporate human input for handling misclassifications and praxis. This paper presents a lung annotation tool for CT images that is interactive, efficient, and robust. The proposed annotation tool produces an "`as accurate as possible'' initial annotation based on the fuzzy-connectedness image segmentation, followed by efficient manual fixation of the initial extraction if deemed necessary by the practitioner. To provide maximum flexibility to the users, our annotation tool is supported in three major operating systems (Windows, Linux, and the Mac OS X). The quantitative results comparing our free software with commercially available lung segmentation tools show higher degree of consistency and precision of our software with a considerable potential to enhance the performance of routine clinical tasks.

\end{abstract}

\section{INTRODUCTION}
Lung related diseases are one of the leading causes of death worldwide. The American Lung Association estimated over 400,000 deaths per year in the United States alone linked to lung diseases \cite{AmericanLungAssociation2013}. For non-invasive diagnosis of lung diseases, computed tomography (CT) is the current standard in routine clinical environment. With the increased exposure of computer-assisted diagnosis methods, automated analysis tools are often sought for quick and accurate diagnosis and quantification \cite{mansoor2014statistical, xu2013559}. Specific to pulmonary diseases, robust and accurate tools are needed to extract information pertaining to the lungs. The delineation of lungs from a CT scan is mostly done manually by the expert in the routine clinics. Although a lot of interest has been shown lately in developing automated methods for lung segmentation such as \cite{armato20041011, kuhnigk2005525, zhang2001204, jones1997113} to name a few, these methods usually fail to gain popularity in practice due to pulmonary abnormalities and also because of tradition and praxis. Additionally, most methods presented in the literature lack an accompanied user-friendly application that can be tested and evaluated by the practitioners since most of these methods require complex parameter adjustments; therefore, it becomes very difficult for a practitioner without enough knowledge of the underlying algorithm. Also most applications allow only the choice of accepting and rejecting the automated quantification; the user in these cases typically does not have the option of correcting the misclassification but have to reject it altogether. Consequently, most current state-of-the-art methods for pulmonary analysis do not gain popularity in the clinical environment. Commercially available software, on the contrary, come with a hefty price tag and usually have very specific hardware and software requirements.

In this paper, we present a robust, fast, and flexible single-click solution to lung-field annotation in CT images. The presented software, CIDI-Lung-Seg, combines manual and automated annotation thus providing what can be considered a computer-aided annotation and quantification. In contrast to completely automated approaches, CIDI-Lung-Seg provides an initial estimation of the lung-field based on region growing-based FC segmentation algorithm, the details of the algorithm can be found in \cite{ciesielski2012375}. The initial annotation can be subsequently modified if deemed necessary by the practitioner. In addition, the software does not require any specific operating system, software, or hardware; since it has been made available in Microsoft Windows, Mac OS X, and linux. To make this manuscript self-contained, a brief outline of the methodology is presented in Section \ref{sec:methods}. Section \ref{sec:software} describes the software, while the performance evaluation with the commercial software tools is presented in Section \ref{sec:performance}. The paper is concluded in Section \ref{sec:conclusion} with a discussion and future directions.

\section{METHODS}
\label{sec:methods}
\begin{figure}[ht]
\centering
\includegraphics[scale =0.45]{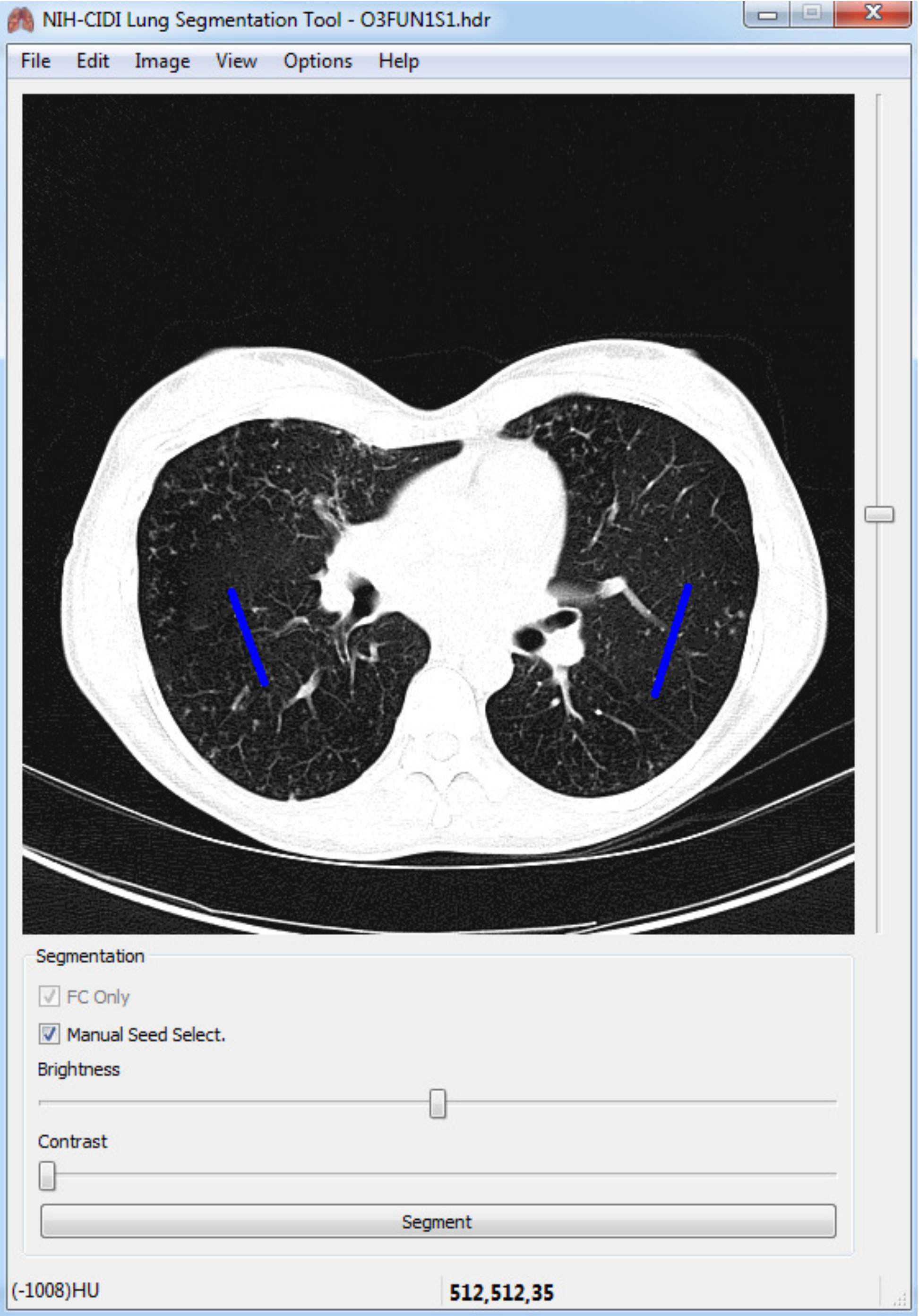}
\caption{Manual seed selection in CIDI-Lung-Seg. Single-stroke painted line (shown in \emph{blue}) are used as seeds for left and right lungs.}
\label{fig:seed_select}
\end{figure}

The algorithm driving the CIDI-Lung-Seg is summarized below; herein, we provide a brief summary of the algorithm. The initial lung parenchyma extraction is performed by adapting the fuzzy-connectedness (FC) image segmentation algorithm \cite{zhou2007348, Udupa1996246}. The initial phase consists of two primary stages: (i) seed selection for FC, and (ii) FC segmentation. The CIDI-Lung-Seg software can identify seed locations automatically as well as it allows the user to manually select them if desired (Fig. \ref{fig:seed_select}). These two steps are explained in the following subsections.

\subsection{Automatic Seed selection and Initial FC segmentation}
\begin{figure}[ht]
\centering
\includegraphics[scale =0.25]{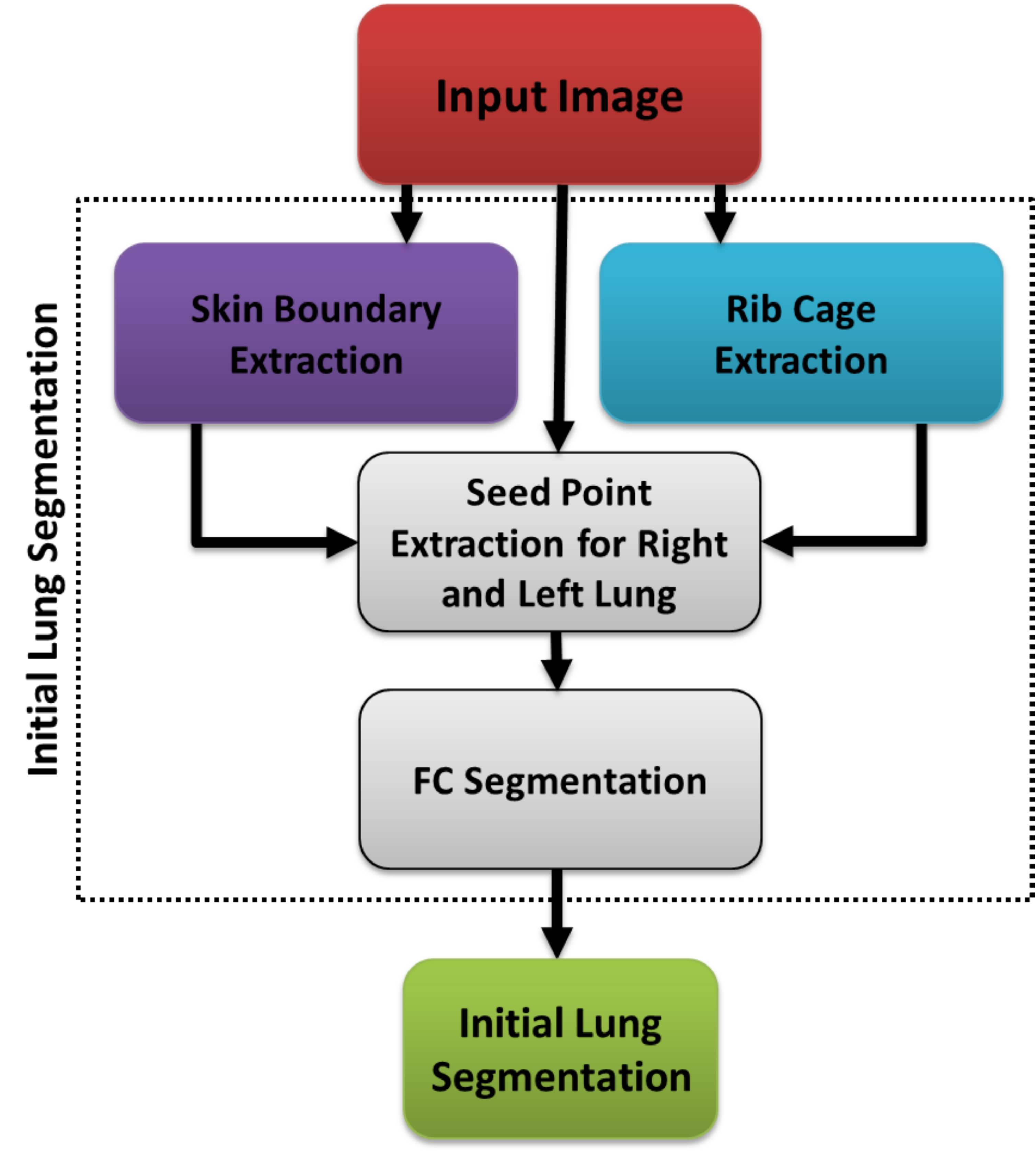}
\caption{Flowchart explaining the initial FC annotation.}
\label{fig:FC_Initial_Seg}
\end{figure}
\begin{figure*}[ht]
\centering
\includegraphics[scale =0.21]{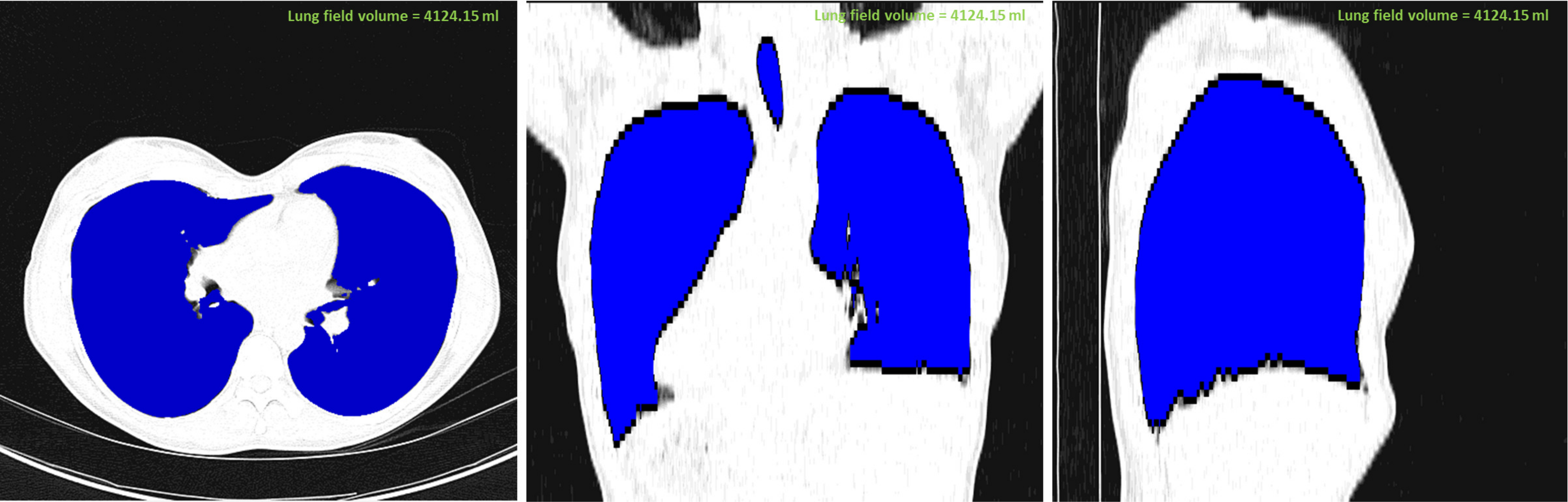}
\caption{Annotated lung (shown in \emph{blue}) from CT image viewed in CIDI-Lung-Seg (a) axial plane, (b) coronal plane, and (c) sagittal plane.}
\label{fig:example_3views}
\end{figure*}
\begin{figure}[ht]
\centering
\includegraphics[scale =0.2]{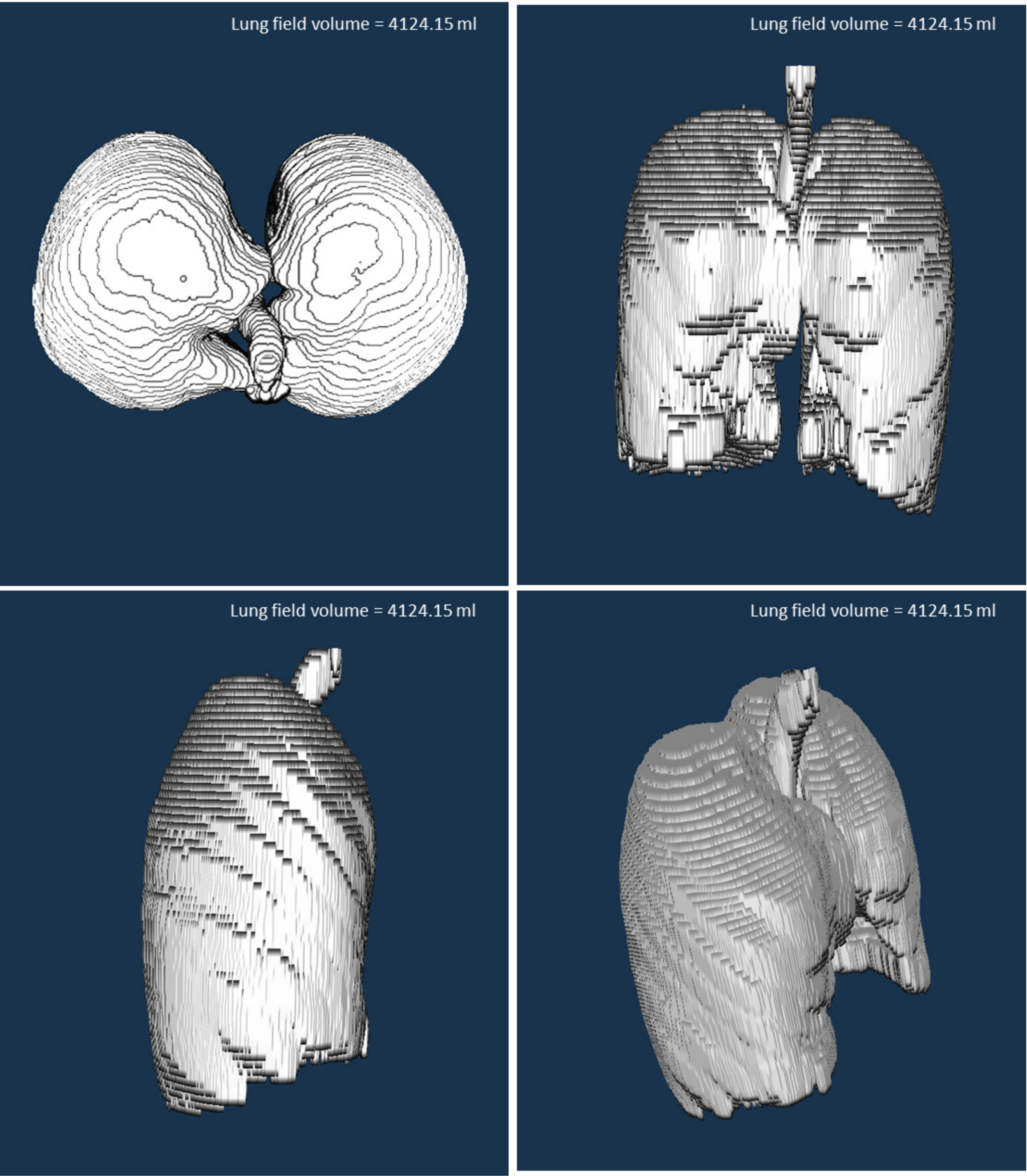}
\caption{3D rendering of the annotation in CIDI-Lung-Seg.}
\label{fig:rendering}
\end{figure}

Fig. \ref{fig:FC_Initial_Seg} summarizes the initial lung annotation process using the FC algorithm. To annotate both lungs, the FC requires two initial seed points $s_l$, and $s_r$, located in the left and right lung parenchyma respectively. As mentioned earlier, these seed points can be manually assigned by users or can be extracted in a fully automated manner. The software assumes the intensity values of the target CT image to be in Hounsfield units (HU). For automatic seed localization, we obtain candidate seeds from a region through a strict thresholding. For a target CT image $I$, the methods started by extracting the geometrical markers, i.e., the skin boundary and rib cage from the image to automatically locate the lung region. The final seed selection is done using the threshold operator $\mathcal{I}$ over CT attenuation values for strictly normal lung parenchyma (HU: -700 through -400, mean $\approx$ -550). Once most robust normal (healthy) regions in left and right lungs are identified, the voxels belonging to the region having the minimum HU values are selected as seed locations $s_l$ and $s_r$ respectively:
\begin{eqnarray}
s_l \leftarrow  \mathcal{L} (\min_{HU} ROI) \in I^{\mathcal{T}}_{left}, \nonumber \\
s_r \leftarrow  \mathcal{L} (\min_{HU} ROI) \in I^{\mathcal{T}}_{right},
\end{eqnarray}
where $\mathcal{L} $ denotes the location of the voxel(s), and $I^{\mathcal{T}} = I^{\mathcal{T}}_{left} \cup I^{\mathcal{T}}_{right}$. In addition to the seeds, the FC algorithm requires the mean $m$ and the variance $\sigma$ of the target region(s). These values are empirically adjusted in the software to the default values corresponding to normal lung parenchyma, i.e., $m=-550$ HU, and $\sigma=150$ HU after analyzing hundreds of CT images from wide variety of sources. Once the seeds are identified and parameters adjusted, FC delineation is performed. It is important to note here that, the entire procedure is performed in the user interface using a single click unless users prefer to input seed location themselves. Fig. \ref{fig:example_3views} shows the annotation produced by CIDI-Lung-Seg in axial, coronal, and sagittal planes. Fig. \ref{fig:rendering} shows 3D rendering of the annotation using the software's rendering module.

\subsection{Refinement}
\begin{figure}[ht]
\centering
\includegraphics[scale =0.3]{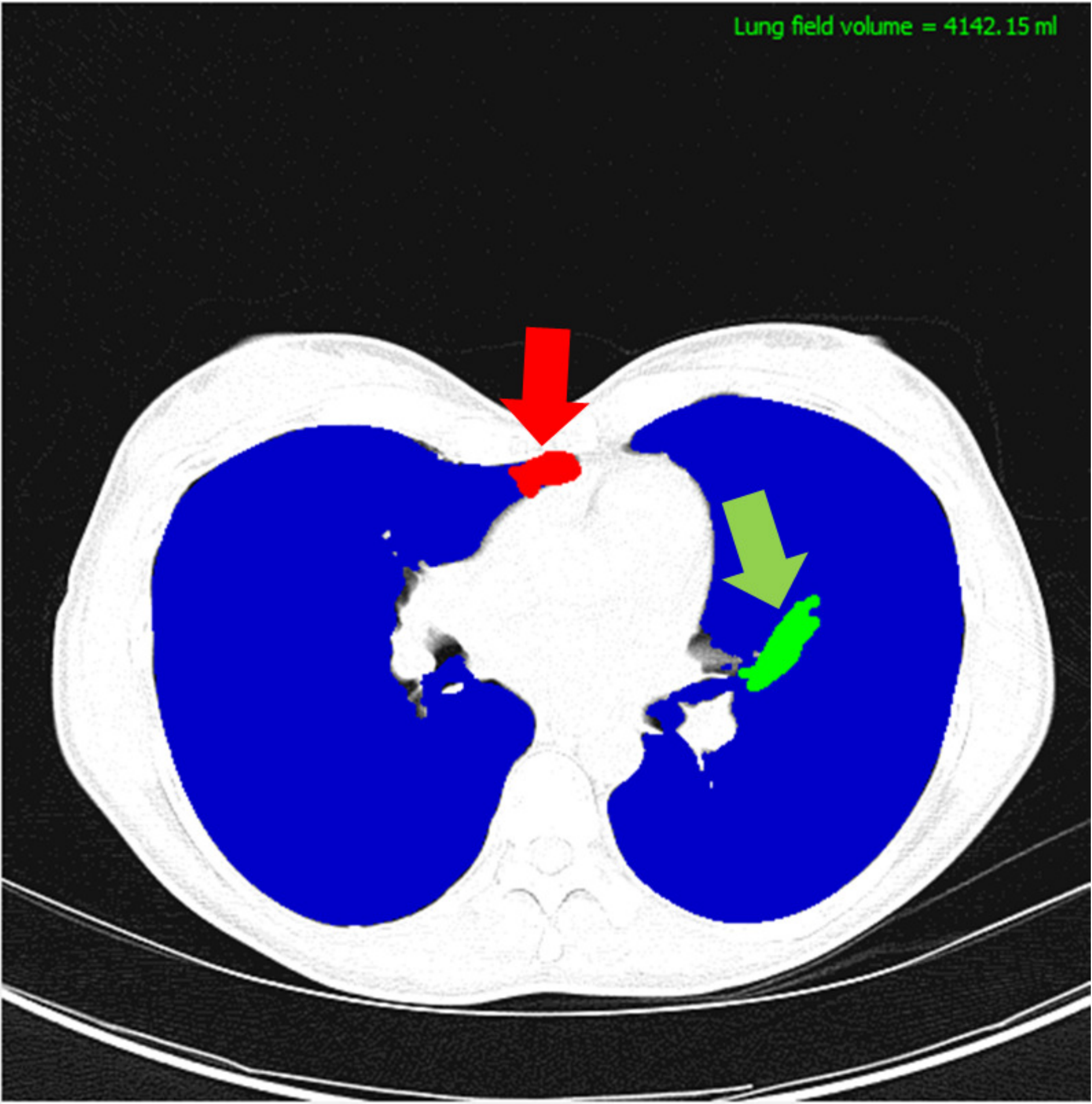}
\caption{Manual correction using the CIDI-Lung-Seg. The practitioner can add (\emph{green}) or delete (\emph{red}) the misclassified areas by stroke painting mechanism.}
\label{fig:refinement}
\end{figure}
Once the initial lung regions are obtained, the annotation results are presented to the user. As shown in the results section, the initial results already provide a normal lung parenchyma for the target with the initial volume estimate. Any refinement (i.e., addition or deletion) in the initial boundaries can be easily incorporated by manual painting tool as shown in Fig. \ref{fig:refinement}. Once the annotation is finalized it can be saved (ANALYZE 7.5 NifTI format) for subsequent analysis. The software currently supports NifTI format, more formats will be supported in the future releases of the software.

\section{Software Description}
\label{sec:software}
CIDI-Lung-Seg is designed and distributed with an open-architecture under the general public GNU license. The software details of CIDI-Lung-Seg is presented in TABLE \ref{table:glossary}. A video guide for the installation and basic working of the software is shared here (\url{http://www.youtube.com/watch?v=VD3GQ0O7weE}). The software uses most common libraries from ITK, VTK, and QT frameworks. Our free software can be downloaded from \url{http://www.nitrc.org/projects/nihlungseg}. 
\begin{table}[!ht]
\renewcommand{\arraystretch}{1.3}
\caption{\small{Overview of the CIDI-Lung-Seg Software.}}
\label{table:glossary}
\centering
\begin{tabular}{lp{5cm}}
\hline
\textbf{Name of Software} & CIDI-Lung-Seg\\
\textbf{Current release} & 1.0.2\\
\textbf{Devlopment} & C/C++, ITK, VTK, QT\\ 
\textbf{Developed at} & Center for Infectious Disease Imaging, National Institutes of Health\\
\textbf{Operating System} & Windows (32-bit, 64-bit), Linux, Mac \\
\textbf{Type} & Medical image analysis\\
\textbf{License} & GNU General Public License\\
\textbf{Input image format} & ANALYZE 7.5 NifTI \\
\textbf{DICOM Support} & In future releases\\
\textbf{Download URL} & \url{http://www.nitrc.org/projects/nihlungseg}\\
\hline
\end{tabular}
\end{table}

\section{Performance Analysis}
\label{sec:performance}
\begin{figure}[ht]
\centering
\includegraphics[scale =0.22]{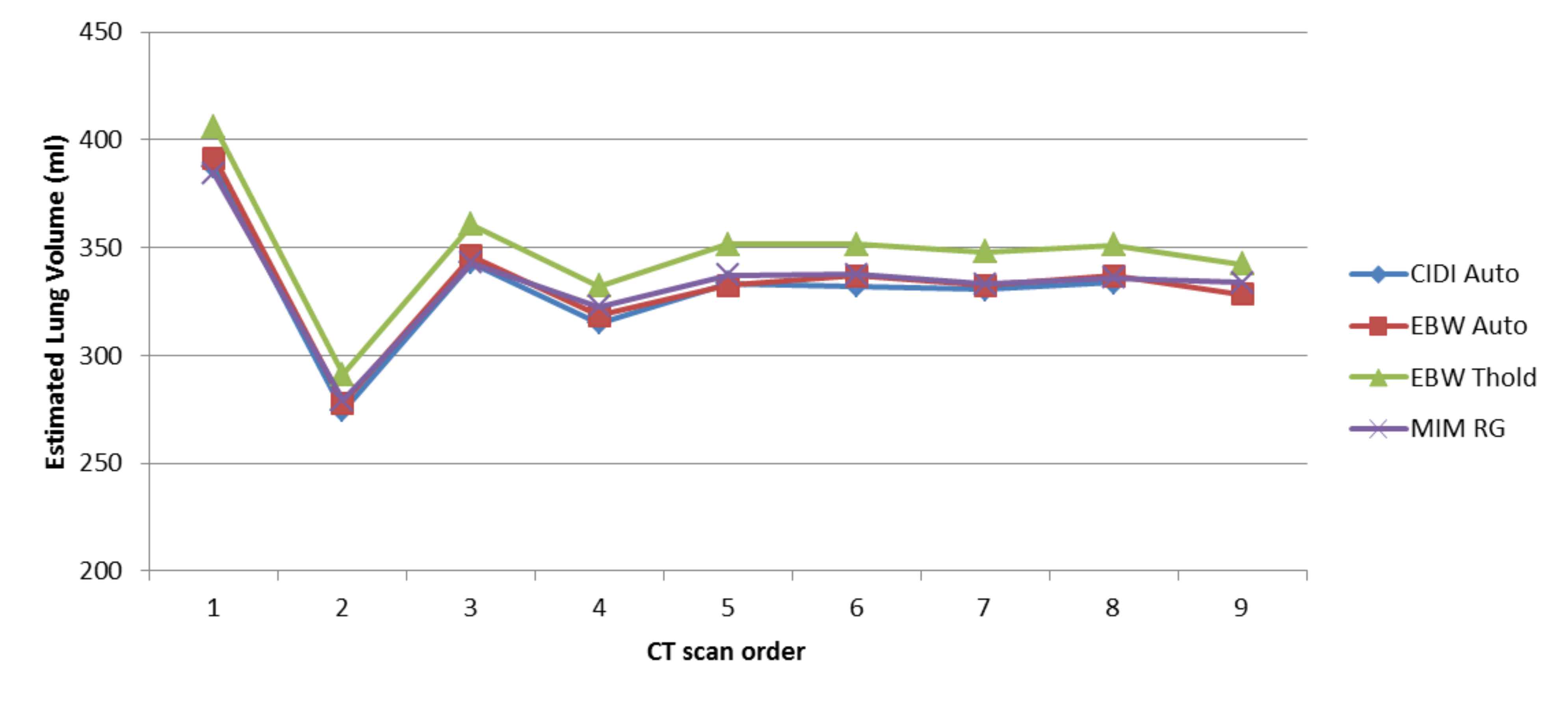}
\caption{Lung volume estimation from animal CT images using different software tools.}
\label{fig:comparison}
\end{figure}
To evaluate the performance of our software against the commercially available tools, we chose two state-of-the-art commercial software tools (i.e., MIM by MIM Software Inc., and Philips Extended Brilliance Workspace (EBW) by Philips Electronics). The segmentation with the MIM Software is performed using its region growing module while Philips EBW is run in two different modes: automated and thresholding, for this comparison. The performance evaluation is performed on 9 in-house collected images at our facility. The estimated lung volume using different approaches is presented in Fig. \ref{fig:comparison}. Table \ref{table:glossary} gives the correlation matrix of the volume for the annotations produced using different software tools. All methods used in the evaluation produced highly correlated results, although it has been noted that MIM region growing miss parts of the lung with HU intensities less than -1000 HU.
\begin{table}[!ht]
\renewcommand{\arraystretch}{1.3}
\caption{\small{Correlation matrix of volumes (mL) for the annotations obtained through different methods.}}
\label{table:glossary}
\centering
\begin{tabular}{p{1.5cm}cp{1cm}p{1.5cm}c}
\hline
&\textbf{CIDI-Lung-Seg} &\textbf{EBW Auto} &\textbf{EBW Threshold} &\textbf{MIM}\\
\hline
\textbf{CIDI-Lung-Seg} & 1 \\
\textbf{EBW Auto} & 0.995 &1 \\
\textbf{EBW Threshold} & 0.999 & 0.996 &1\\
\textbf{MIM} & 0.996 & 0.990 & 0.995 &1\\
\hline
\end{tabular}
\end{table}

\subsection{Independent external evaluation}
To analyze the effectiveness of our software and for an independent comparison, we tested CIDI-Lung-Seg using publically available LObe and Lung Analysis 2011 (LOLA11) Challenge data set \url{http://lola11.com/}. The challenge data consists of $55$ anonymized CT scans, the abnormality in the scans ranges from mild to severe. Results submitted to the challenge organizers were evaluated against a reference standard using overlap coefficient and published online. The results were reported in terms of minimum, mean, median, and maximum overlap over the 55 scans. The final score was the mean over all scans. The evaluation provided for CIDI-Lung-Seg is reproduced in Table \ref{table:lola}.
\begin{table}[ht]
\caption{Overlap coefficient for the CIDI-Lung-Seg for the 55 scans in LOLA11 challenge.}\label{table:lola}
		\begin{tabular}{p{1cm}|r r r r r r r}
		\hline
		obj & mean & std & min & Q1 & median & Q3 & max \\
		\hline
		Left lung & 0.968&	0.097&	0.316&	0.979&	0.987&	0.995&	0.999\\
		Right lung &0.968	&0.134&	0.000&	0.984&	0.990&	0.997&	0.999\\
		\hline
		score & 0.968 & & & & & &\\ 
		\hline
		\end{tabular}\\	
\end{table}

\section{Conclusion and Future Directions}
\label{sec:conclusion}
In this paper, we presented an open source, multi-platform, single-click lung extraction from CT images. The method takes a leap from traditional manual annotation software tools; at the same time it avoids simplistic fully automated approaches that allow clinicians to completely accept or reject annotations these methods produce. The software provides an automated initial annotation, the software provides an easy to use paint tool to correct the initial guess. The comparative tests performed with commercially available software tools demonstrate the robustness and reliability of our software.

In the future releases of the software, we plan to incorporate fast, robust, and efficient machine-learning methods for accurate pathology annotation and quantification. In addition, automatic techniques to segment trachea, airway wall \cite{xu2013113701}, and lung-lobes are envisioned in the future releases of CIDI-Lung-Seg. The inclusion of these methods will make the CIDI-Lung-Seg a complete toolkit for comprehensive pulmonary analysis that can aide clinicians in better diagnosis.

\bibliographystyle{IEEEtran}
\bibliography{IEEELung} 

\begin{thebibliography}{10}
\providecommand{\url}[1]{#1}
\csname url@samestyle\endcsname
\providecommand{\newblock}{\relax}
\providecommand{\bibinfo}[2]{#2}
\providecommand{\BIBentrySTDinterwordspacing}{\spaceskip=0pt\relax}
\providecommand{\BIBentryALTinterwordstretchfactor}{4}
\providecommand{\BIBentryALTinterwordspacing}{\spaceskip=\fontdimen2\font plus
\BIBentryALTinterwordstretchfactor\fontdimen3\font minus
  \fontdimen4\font\relax}
\providecommand{\BIBforeignlanguage}[2]{{%
\expandafter\ifx\csname l@#1\endcsname\relax
\typeout{** WARNING: IEEEtran.bst: No hyphenation pattern has been}%
\typeout{** loaded for the language `#1'. Using the pattern for}%
\typeout{** the default language instead.}%
\else
\language=\csname l@#1\endcsname
\fi
#2}}
\providecommand{\BIBdecl}{\relax}
\BIBdecl

\bibitem{AmericanLungAssociation2013}
``Estimated prevalence and incidence of lung disease,'' American Lung
  Association, Washington, DC, Tech. Rep., April 2013.

\bibitem{mansoor2014statistical}
A.~Mansoor, V.~Patsekin, D.~Scherl, J.~Robinson, and B.~Rajwa, ``A statistical
  modeling approach to computer-aided quantification of dental biofilm,''
  \emph{Biomedical and Health Informatics, IEEE Journal of}, pp. 1--1, 2014.

\bibitem{xu2013559}
Z.~Xu, U.~Bagci, B.~Foster, A.~Mansoor, and D.~J. Mollura, ``Spatially
  constrained random walk approach for accurate estimation of airway wall
  surfaces,'' in \emph{Medical Image Computing and Computer-Assisted
  Intervention--MICCAI 2013}.\hskip 1em plus 0.5em minus 0.4em\relax Springer,
  2013, pp. 559--566.

\bibitem{armato20041011}
S.~G. Armato~III and W.~F. Sensakovic, ``Automated lung segmentation for
  thoracic {CT}: Impact on computer-aided diagnosis,'' \emph{Academic
  Radiology}, vol.~11, no.~9, pp. 1011--1021, 2004.

\bibitem{kuhnigk2005525}
J.-M. Kuhnigk, V.~Dicken, S.~Zidowitz, L.~Bornemann, B.~Kuemmerlen, S.~Krass,
  H.-O. Peitgen, S.~Yuval, H.-H. Jend, W.~S. Rau \emph{et~al.}, ``New tools for
  computer assistance in thoracic {CT}. part 1. functional analysis of lungs,
  lung lobes, and bronchopulmonary segments,'' \emph{Radiographics}, vol.~25,
  no.~2, pp. 525--536, 2005.

\bibitem{zhang2001204}
L.~Zhang, E.~A. Hoffman, and J.~M. Reinhardt, ``Lung lobe segmentation by graph
  search with 3d shape constraints,'' in \emph{Proc. SPIE}, vol. 4321, 2001,
  pp. 204--215.

\bibitem{jones1997113}
T.~N. Jones and D.~N. Metaxas, ``Automated 3d segmentation using deformable
  models and fuzzy affinity,'' in \emph{Information Processing in Medical
  Imaging}.\hskip 1em plus 0.5em minus 0.4em\relax Springer, 1997, pp.
  113--126.

\bibitem{ciesielski2012375}
K.~C. Ciesielski, J.~K. Udupa, A.~X. Falc{\~a}o, and P.~A. Miranda, ``Fuzzy
  connectedness image segmentation in graph cut formulation: A linear-time
  algorithm and a comparative analysis,'' \emph{Journal of Mathematical Imaging
  and Vision}, vol.~44, no.~3, pp. 375--398, 2012.

\bibitem{zhou2007348}
Y.~Zhou and J.~Bai, ``Multiple abdominal organ segmentation: An atlas-based
  fuzzy connectedness approach,'' \emph{Information Technology in Biomedicine,
  IEEE Transactions on}, vol.~11, no.~3, pp. 348--352, 2007.

\bibitem{Udupa1996246}
J.~K. Udupa and S.~Samarasekera, ``Fuzzy connectedness and object definition:
  Theory, algorithms, and applications in image segmentation,'' \emph{Graphical
  Models and Image Processing}, vol.~58, no.~3, pp. 246 -- 261, 1996.

\bibitem{xu2013113701}
Z.~Xu, U.~Bagci, A.~Kubler, B.~Luna, S.~Jain, W.~R. Bishai, and D.~J. Mollura,
  ``Computer-aided detection and quantification of cavitary tuberculosis from
  ct scans,'' \emph{Medical physics}, vol.~40, no.~11, p. 113701, 2013.

\end{thebibliography}

\end{document}